\documentclass[11pt,a4paper]{article}
\pdfoutput=1
\usepackage{coling2016}
\usepackage[utf8]{inputenc}
\usepackage{times}
\usepackage{url}
\usepackage{latexsym}
\usepackage{amsmath}
\usepackage{amsfonts}
\usepackage{graphicx}
\usepackage{tipa}
\usepackage{sidecap}
\usepackage{graphicx}

\title{From phonemes to images: levels of representation in a recurrent neural
model of visually-grounded language learning}
  \author{Lieke Gelderloos \\
    Tilburg University \\
    {\tt liekegelderloos@gmail.com} \And
    Grzegorz Chrupała \\
    Tilburg University \\
    {\tt g.chrupala@uvt.nl} }

\date{}

\begin{document}
\maketitle
\begin{abstract}
We present a model of visually-grounded language learning based on stacked gated recurrent neural networks which learns to predict visual features given an image description in the form of a sequence of phonemes. The learning task resembles that faced by human language learners who need to discover both structure and meaning from noisy and ambiguous data across modalities. We show that our model indeed learns to predict features of the visual context given phonetically transcribed image descriptions, and show that it represents linguistic information in a hierarchy of levels: lower layers in the stack are comparatively more sensitive to form, whereas higher layers are more sensitive to meaning.
\end{abstract}

\blfootnote{
	\hspace{-0.65cm}  
	This work is licensed under a Creative Commons 
	Attribution 4.0 International License.
	License details:
	\url{http://creativecommons.org/licenses/by/4.0/}
}

\section{Introduction}
\label{sec:intro}
Children acquire their native language with little and weak
supervision, exploiting noisy correlations between speech, visual, and
other sensory signals, as well as via feedback from interaction with
their peers and parents. Understanding this process is an important
scientific challenge in its own right, but it also has potential to
generate insights useful in engineering efforts to design
conversational agents or robots. Computationally modeling the ability
to learn linguistic form--meaning pairings has been the focus of much
research, under scenarios simplified in a variety of ways, for
example:

\begin{itemize}
\setlength\itemsep{-0.5em}
\item distributional learning from pure word-word co-occurrences with
  no perceptual grounding \cite{landauer1998introduction,kiros2015skip};
\item cross-situational learning with word sequences and sets of
  symbols representing sensory input
  \cite{siskind.96,fazly.etal.10csj};
\item cross-situational learning using sensory audio and visual
  input, but with extremely limited sets of words and objects
  \cite{Roy2002113,iwahashi2003language}.
\end{itemize}

Some recent models have used more naturalistic, larger-scale inputs,
for example in cross-modal distributional semantics
\cite{lazaridou2015combining} or in implementations of the acquisition
process trained on images paired with their descriptions
\cite{chrupala2015learning}. While in these works the representation
of the {\it visual scene} consists of pixel-level perceptual data, the
{\it linguistic} input consists of sentences segmented into discrete
word symbols. In this paper we take a step towards addressing this
major limitation, by using the phonetic transcription of input
utterances. While this type of input is symbolic rather than
perceptual, it goes a long way toward making the setting more
naturalistic, and the acquisition problem more challenging: the
learner may need to discover structure corresponding to morphemes, words
and phrases in an unsegmented string of phonemes, and the length of
the dependencies that need to be detected grows substantially when
compared to word-level models.

\paragraph{Our contributions}

We design and implement a simple architecture based on stacked
recurrent neural networks with Gated Recurrent Units
\cite{chung2014empirical}: our model processes the utterance phoneme
by phoneme while building a distributed low-dimensional semantic
representation through a series of recurrent layers.  The model learns
by projecting its sentence representation to image
space and comparing it to the features of the corresponding visual
scene.

We train this model on a phonetically transcribed
version of MS-COCO \cite{lin2014microsoft} and show that it is able to
successfully learn to understand aspects of sentence meaning from the
visual signal, and exploits temporal structure in the input. In a
number of experiments we show that different levels in the stack of
recurrent layers represent different aspects of linguistic
structure. Low levels focus on local, short-time-scale spans of the
input sequence, and are comparatively more sensitive to form. The top
level encodes global aspects of the input sequence and is sensitive to
visually salient elements of its meaning.

\section{Related work}
A major part of learning language consists in learning its structure,
but in order to be able to communicate it is also of crucial importance
to learn the relation of words to entities in the
world.
Grounded lexical acquisition is often modeled as cross-situational
learning, a process of rule-based \cite{siskind.96} or statistical
\cite{fazly.etal.10csj,frank.etal.07} inference of word-to-referent
mappings.
Cross-situational models typically work on word-level language input
and symbolic representations of the context. However, infants have to
learn from continuous perceptual input. 
\newcite{lazaridou2016multimodal} partially remedy this shortcoming
and propose a model of learning word
meanings from text paired with continuous image representations; the
limitation of their work is the toy evaluation dataset.

Recent experimental and computational studies have found that co-occurring visual information may help to learn word forms \cite{thiessen2010effects,Cunillera2010295,Glicksohn2013,Yurofsky2012statistical}. This suggests that acquisition of word form and meaning are interactive, rather than separate.

The Cross-channel Early Lexical Learning (CELL) model of \newcite{Roy2002113} and the more recent work of \newcite{rasanen2015joint} take into account the continuous nature of the speech signal, and incorporate visual information as well. The CELL model learns to discover words in continuous speech through co-occurence with their visual referent, but the visual input only consists of the shape of single objects, effectively bypassing referential uncertainty. \newcite{rasanen2015joint} propose a probabilistic joint model of word segmentation and meaning acquisition from raw speech and a set of possible referents that appear in the context. In both \newcite{Roy2002113} and \newcite{rasanen2015joint} the visual context is considerably less noisy and ambiguous than that available to children.

There is an extensive line of research on image captioning (see \newcite{bernardi2016automatic} for a recent overview). Typically, captioning models learn to recognize high-level image features and associate them with words. Inspired by both image captioning research and cross-situational human language acquisition, two recent automatic speech recognition models learn to recognize word forms from visual data. In \newcite{synnaeve2014learning}, language input consists of single spoken words and visual data consists of image fragments, which the model learns to associate. \newcite{harwath2015deep} employ two convolutional neural networks, a visual object recognition model and a word recognition model, and an embedding alignment model that learns to map recognized words and objects into the same high-dimensional space. Although the object recognition works on the raw visual input, the speech signal is segmented into words before presenting it to the word recognition model. As both \newcite{harwath2015deep} and \newcite{synnaeve2014learning} work with word-sized chunks of speech, they bypass the segmentation problem that human language learners face.

Character-level input representations have recently gained attention
in NLP. \newcite{ling2015finding} and
\newcite{plank2016multilingual} use bidirectional LSTMs to compose
characters into word embeddings, while
\newcite{chung2016character} propose machine translation model with 
character level output. These approaches exploit character-level
information but crucially they assume that word boundaries are available
in the input.

Character-level neural NLP \textit{without} explicit word boundaries in the input is studied in cases where fixed vocabularies are inherently problematic, e.g.\ with combined natural and programming language input \cite{chrupala2013text} or when specifically dealing with misspelled words in automatic writing feedback \cite{xie2016neural}. 

Character-level language models are analyzed in
\newcite{hermans2013training} and
\newcite{karpathy2015visualizing}. Both studies show that
character-level recurrent neural networks are sensitive to
long-range dependencies: for example by keeping track of opening and
closing parentheses over stretches of
text. \newcite{hermans2013training} describe the hierarchical
organization that emerges during training, with higher layers
processing information over longer timescales. In our work we show
related effects in a model of visually-grounded
language learning from unsegmented phonemic strings.

We use phonetic transcription of full sentences as a first step
towards large-scale multimodal language learning from speech
co-occurring with visual scenes. 
The use of phonetic
transcription rather than raw speech signal simplifies learning
and allows us to perform experiments on the encoding of linguistic
knowledge as reported in section \ref{sec:experiments} without
additional annotation. Our goal is to model a multi-modal language learning process that includes the segmentation problem faced by human language learners. In contrast to character-level NLP and language modeling approaches, our input data therefore does not contain word boundaries or strong cues such as whitespace and punctuation.

In contrast to \newcite{Roy2002113}
and \newcite{rasanen2015joint}, the visual input to our model consists
of high-level visual features, which means it contains ambiguity and
noise. In contrast to \newcite{synnaeve2014learning} and
\newcite{harwath2015deep}, we consider full utterances rather than
separate words. 

To our knowledge, there is no work yet on multimodal phoneme or character-level language modeling with visual input.
Although the language input in this study is low-level-symbolic rather
than perceptual, the learning problem is similar to that of a human language learner: discover language structure as well as meaning, based on ambiguous and noisy data from another modality. 

\newcite{chrupala2015learning} simulate visually grounded human
language learning in face of noise and ambiguity in the visual
domain. Their model predicts visual context given a sequence of words. While the visual input consists of a continuous representation, the language input consists of a sequence of words. The aim of this study is to take their approach one step further towards multimodal language learning from raw perceptual input. 
\newcite{Kdr2016RepresentationOL} develop techniques for understanding
and interpretation of the representations of linguistic form and
meaning in recurrent neural networks, and apply these to word-level
models. In our work we share the goal of revealing the nature of
emerging representations, but we do not assume words as their basic
unit. Also, we are especially concerned with the emergence of a
hierarchy of levels of representations in stacked recurrent networks. 
\section{Models}
Consider a learner who sees a person pointing at a scene and uttering
the unfamiliar phrase
{\it Look, the monkeys're playing}. We may suppose that the learner
will update her language understanding model such that the
subsequent utterance of this phrase will evoke in her mind something
close to the impression of this visual scene. Our model is a particular
instantiation of this simple idea.
\subsection{Phon GRU}
\begin{figure}
\begin{minipage}[l]{0.45\textwidth}
  \includegraphics[scale=0.2]{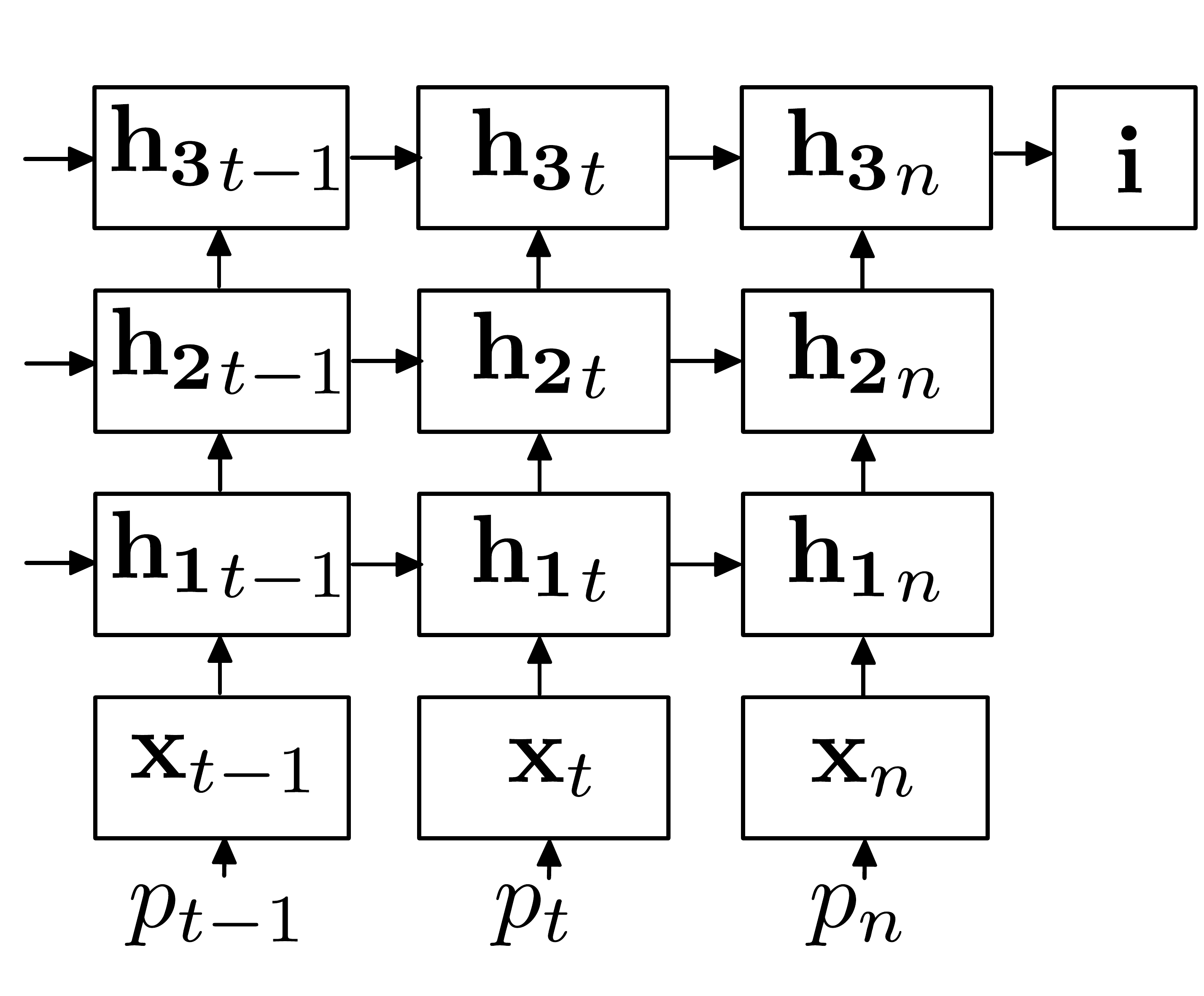}
  \caption{A three-timestep slice of the stacked recurrent architecture with three hidden layers.}
  \label{fig:architecture}
\end{minipage}
\hspace{0.3cm}
\begin{minipage}[r]{0.45\textwidth}
  \begin{tabular}{|l|}\hline
    \includegraphics[scale=0.7]{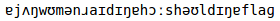} \\
    A young woman riding a horse holding a flag\\\hline
  \end{tabular}
  \begin{center}
    \includegraphics[scale=0.2]{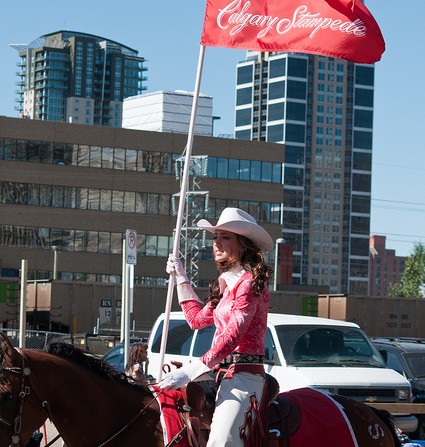}
  \end{center}

  \caption{(Top) Example of a postprocessed phonetic transcription
    from eSpeak used as input to the {\sc Phon GRU} model. (Bottom)
    Corresponding image.}
  \label{fig:ipa}
\end{minipage}
\end{figure}

The architecture of our main model of interest, {\sc Phon GRU} is
schematically depicted in Figure \ref{fig:architecture} and consists of a phoneme
encoding  layer, followed by a stack of $K$ Gated Recurrent Neural
nets, followed by a densely connected layer which maps the last hidden
state of the top recurrent layer to a vector of visual features.

Gated Recurrent Units (GRU) were introduced in
\newcite{cho2014properties} and \newcite{chung2014empirical} as an
attempt to alleviate the problem of vanishing gradient in standard
simple recurrent nets as known since the work of
\newcite{elman1990finding}. GRUs have a linear shortcut through
timesteps which bypasses the nonlinearity and thus promotes gradient
flow.
Specifically, a GRU computes the hidden state at current time step, $\mathbf{h}_{t}$, as the
linear combination of previous activation $\mathbf{h_{t-1}}$, and a new
{\it candidate} activation $\mathbf{\tilde{h}}_t$:

\begin{equation}
  \mathrm{gru}(\mathbf{x}_t, \mathbf{h}_{t-1}) = (1 - \mathbf{z}_t)\odot \mathbf{h}_{t-1} + \mathbf{z}_t \odot \mathbf{\tilde{h}}_t
\vspace{-.1cm}
\end{equation}
where $\odot$ is elementwise multiplication, and the update gate
activation $\mathbf{z_{t}}$ determines the amount of new information
mixed in the current state:

\begin{equation}
\label{eq:gru-update}
   \mathbf{z}_t = \sigma_s(\mathbf{W}_z \mathbf{x}_t + \mathbf{U}_z \mathbf{h}_{t-1})
\end{equation}
The candidate activation is computed as:
\begin{equation}
\label{eq:gru-cand}
   \mathbf{\tilde{h}}_t = \sigma(\mathbf{W} \mathbf{x}_t + \mathbf{U}(\mathbf{r}_t \odot \mathbf{h}_{t-1}))
\end{equation}
The reset gate $\mathbf{r_{t}}$ determines how much of the current
input $\mathbf{x_{t}}$ is mixed in the previous state
$\mathbf{h}_{t-1}$ to form the candidate activation:
\begin{equation}
\label{eq:gru-reset}
   \mathbf{r}_t = \sigma_s(\mathbf{W}_r \mathbf{x}_t + \mathbf{U}_r \mathbf{h}_{t-1})
\end{equation}

By applying the $\mathrm{gru}$ function repeatedly a GRU layer maps a
sequence of inputs to a sequence of states:
\begin{equation}
  \mathrm{GRU}(\mathbf{X}, \mathbf{h}_0) = \mathrm{gru}(\mathbf{x}_n, \dots, \mathrm{gru}(\mathbf{x}_2, \mathrm{gru}(\mathbf{x}_1, \mathbf{h}_0)))
\end{equation}
where $\mathbf{X}$ stands for the matrix composed of input column vectors
$\mathbf{x}_1, \ldots, \mathbf{x}_n$. Two or more GRU layers can be composed into a stack: 
\begin{equation}
\mathrm{GRU}_2(\mathrm{GRU}_1(\mathbf{X}, {\mathbf{h_1}}_{0}), {\mathbf{h_2}}_{0}).
\end{equation}
In our version of the Stacked GRU architecture we use {\it residualized} layers:
\begin{equation}
\mathrm{GRU_{res}}(\mathbf{X}, \mathbf{h}_0) = \mathrm{GRU}(\mathbf{X}, \mathbf{h}_0) + \mathbf{X}
\end{equation}
Residual convolutional networks were introduced by
\newcite{he2015deep}, while \newcite{oord2016pixel} showed their
applicability to recurrent architectures. We adopt residualized layers
here as we observed they speed up learning in stacks of several
GRU layers.

Our gated recurrent units use steep sigmoids for gate activations: \[
\sigma_s(z) = \frac{1}{1 + \exp(-3.75z)} 
\]
and rectified linear units clipped between 0 and 5 for the unit
activations:
\[
\sigma(z) = \mathrm{clip(0.5(z+\mathrm{abs}(z)), 0, 5)}
\]

There are two more components of our {\sc Phon GRU} model: the
phoneme encoding layer, and mapping from the final state of the top GRU
layer to the image feature vector.
The phoneme encoding layer is a simply a lookup table $\mathbf{E}$ whose
columns correspond to one-hot-encoded phoneme vectors. The input
phoneme $p_t$ of utterance $p$ at each step $t$ indexes into the
encoding matrix and produces the input column vector:
\begin{equation}
  \mathbf{x}_t = \mathbf{E}[:,p_t].
\end{equation}
Finally, we map the final state of the top GRU layer ${\mathbf{h_K}}_n$
to the vector of image features using a fully connected layer:

\begin{equation}
  \hat{\mathbf{i}} = \mathbf{I} {\mathbf{h_K}}_n
\end{equation}

Our main interest lies in recurrent phoneme-level modeling. However, in order to
put the performance of the phoneme-level {\sc Phon GRU} into
perspective, we compare it to two word-level models. Importantly,
the word models should {\bf not} be seen as baselines, as they have access to
word boundary and word identity information not available to
{\sc Phon GRU}. 

\subsection{Word GRU}
The architecture of this model is the same as {\sc Phon GRU} with
the difference that we use words instead of phonemes as input symbols,
use learnable word embeddings instead of fixed one-hot phoneme
encodings, and reduce the number of layers in the GRU stack. See
Section~\ref{sec:experiments} for details.
\subsection{Word Sum}
The second model we use for comparison is a word-based non-sequential
model, consisting of a word embedding matrix, a vector sum operator,
and a mapping to the image feature vector:
\begin{equation}
  \label{eq:sum}
  \hat{\mathbf{i}} = \mathbf{I} \sum_{t=1}^n \mathbf{E}[:,w_t]
\end{equation}
where $w_t$ is the word at position $t$ in the input utterance.
This model simply learns word embeddings which are then summed into a
single vector and projected to the target image vector. Thus this model does
not have access to word sequence information, and is a distributed
analog of a bag-of-words model.

\section{Experiments}
\label{sec:experiments}

For all experiments, the models were trained on the training set of MS-COCO. MS-COCO contains over 163,000 images accompanied by at least five captions by human annotators. With an average of 7.7 labeled object instances per image, images typically contain more objects than the captions mention, making reference to the scene ambiguous. Textual input for the {\sc Phon GRU} models was transcribed automatically using the grapheme-to-phoneme functionality with the default English voice of the eSpeak speech synthesis toolkit.\footnote{Available at \url{http://espeak.sourceforge.net}} Stress and pause markers were removed, as well as word boundaries (after storing their position for use in experiments), leaving only phoneme symbols. See Figure~\ref{fig:ipa} for an example transcription.

Visual input for all models was obtained by forwarding images through the 16-layer convolutional neural network described in \newcite{simonyan2014very} pre-trained on Imagenet \cite{ILSVRCarxiv14}, and recording the activation vectors of the pre-softmax layer. The z-score transformation was applied to these features to ease optimization. 

Most of the details of the three model types and training
hyperparameters were adopted from related work, and adapted via
a limited amount of exploratory experimentation. Exhaustive exploration of the search space was not feasible due to the large number of adjustable settings in these models and their long running time. Given the importance of depth for our purposes, we did systematically explore the number of layers in the {\sc Phon GRU} and {\sc Word GRU} models. A single layer is optimal for {\sc Word GRU}. For {\sc Phon GRU}, see Section~\ref{subsec:visual} below. Other important settings were as follows:
\begin{itemize}
\setlength\itemsep{-0.5em}
\item All models: Implemented in Theano \cite{Bastien-Theano-2012}, optimized with 
  Adam \cite{DBLP:journals/corr/KingmaB14}, initial learning rate of 0.0002, minibatch size
  of 64, gradient norm clipped to 5.0.
\item {\sc Word Sum}: 1024-dimensional word embeddings, words with frequencies below 10 replaced by {\tt UNK} token.
\item {\sc Word GRU}: 1024-dimensional word embeddings, a single 1024 dimensional hidden layer, words with frequencies below 10 replaced by {\tt UNK} token.
\item {\sc Phon GRU}: 1024-dimensional hidden layers.
\end{itemize}

\subsection{Prediction of visual features}
\label{subsec:visual}
The models are trained and evaluated on the prediction of visual
feature vectors from captions. While our goal is not to develop an
image retrieval method, we use this task as it reflects the ability to extract visually salient semantic information from language.
For the experiments on the prediction of visual features all models
were trained on the training set of MS-COCO. As validation and test data we
used a random sample of 5000 images each from the MS-COCO validation set. 

\begin{figure}
    \centering
  \begin{minipage}{0.48\textwidth}
    \includegraphics[scale=0.32]{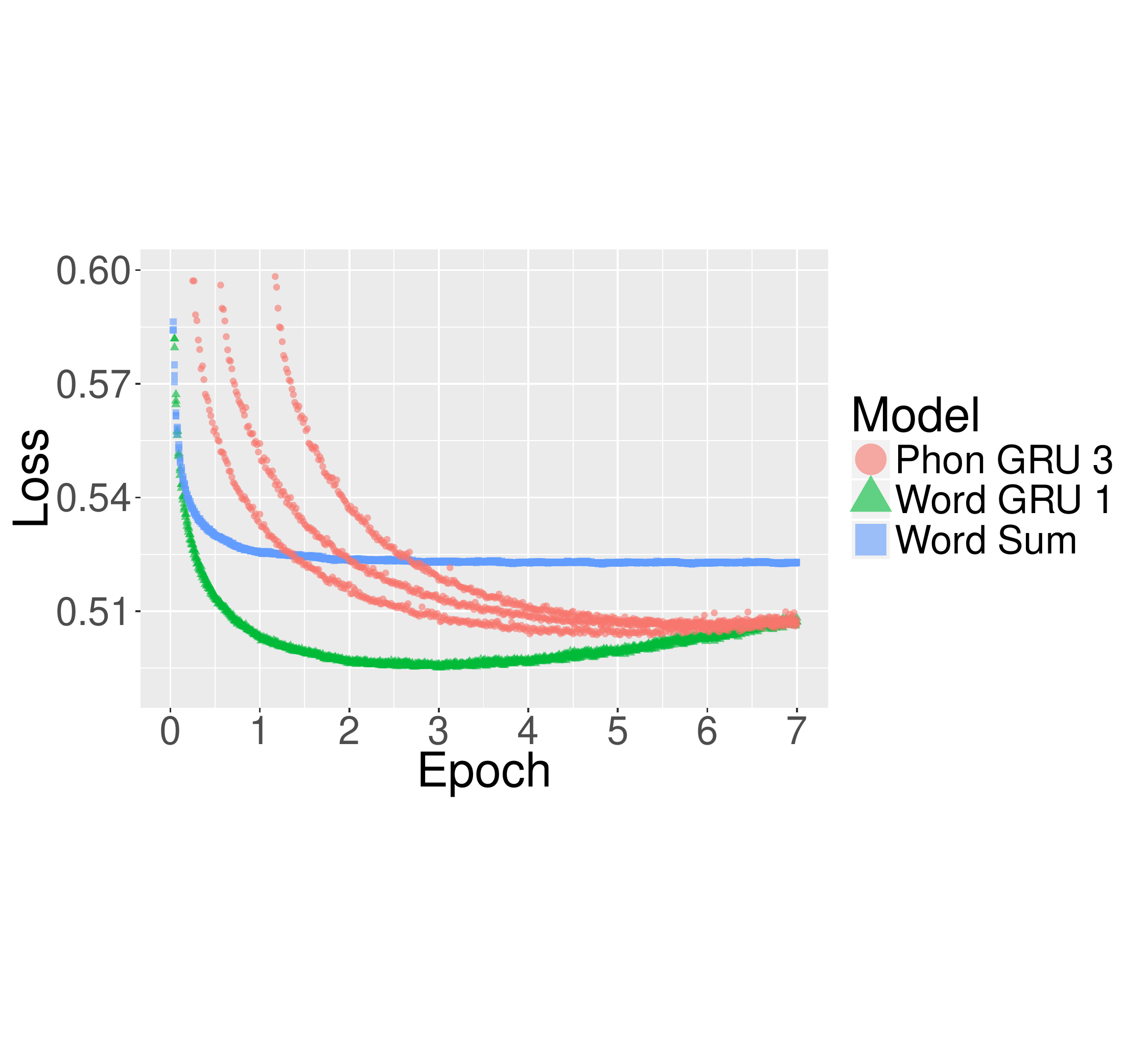}
    \caption{Value of the loss function on validation data during
      training. Three random initialization of each model are shown.}
    \label{fig:loss}
  \end{minipage}
\hspace{0.3cm}
  \begin{minipage}{0.48\textwidth}
    \includegraphics[scale=0.32]{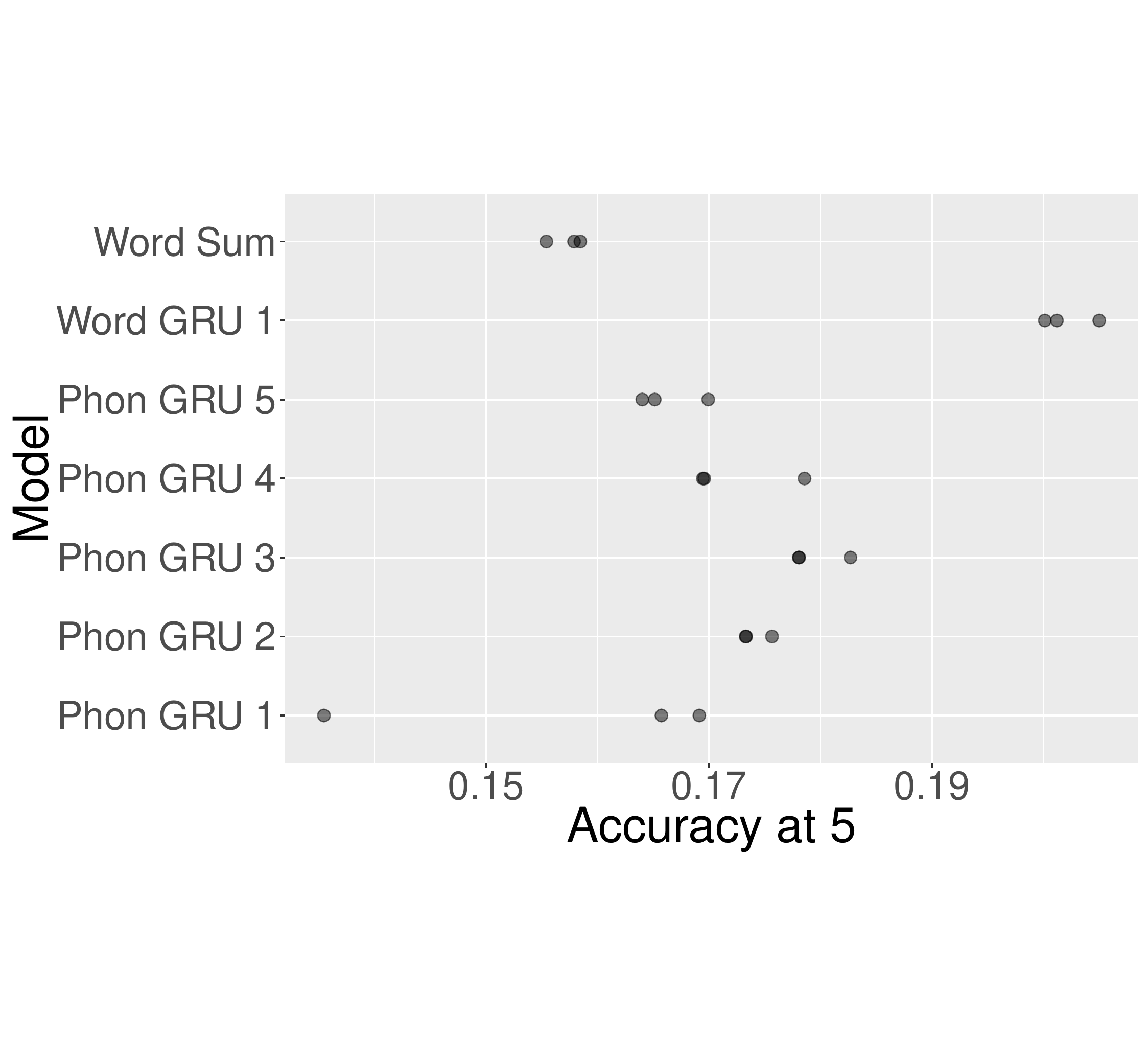}
    \caption{Validation accuracy at 5 on the image retrieval task.}
    \label{fig:accat5}
  \end{minipage}
\end{figure}

Figure~\ref{fig:loss} shows the value of the validation average cosine distance
between the predicted visual vector and the target vector for three
random initializations of each of the model types. 

The Phonetic GRU model is more sensitive to the initialization: one
can clearly distinguish three separate trajectories. The word-level models
are much less affected by random initialization. In terms of the
overall performance, the {\sc Phon GRU} model falls between the
{\sc Word Sum} model and the {\sc Word GRU} model.

We also evaluated the models on how well they perform when used to
search images: for each validation sentence the model was used to predict the
visual vector. The image vectors in the validation data were then
ranked by cosine similarity to the predicted vector, and the
proportion of times the correct image was among the top 5 was
reported. By {\it correct} image we mean the one which the sentence
was used to describe (even though often many other images are also
good matches to the sentence). 

In Figure~\ref{fig:accat5} we report the validation accuracies on this
task for the two word-level models, as well as for the Phon GRU
model with different number of hidden layers. We trained each model
version with three random initializations for each model setting, and
evaluate after each epoch. We report the score of the best epoch for
each initialization. 
The overall ranking of the models matches the direct
evaluation of the loss function above: the phoneme-level models are in
between the two word-level models. {\sc Phon GRU} with three
hidden layers is the best of the phoneme-level models.

In Table~\ref{tab:accat5test} we show the accuracies of the best
version of each of the models types on the test images; these are also
the model versions used in all subsequent experiments. The scores for 
the {\sc Word GRU} are comparable to what
\newcite{chrupala2015learning} report for their multitask {\sc
  Imaginet} model, whose visual pathway has the same
structure, and who use the same data. 
More recently, \newcite{vendrov2015order} report substantially higher
scores for image search with a word-level GRU model, with the
following main differences from our setting: better image features,
larger training set, and a loss function optimized for the ranking
task.\footnote{We have preliminary results indicating that most of the
  analyses in the rest of Section~\ref{sec:experiments} show the same general
  pattern for phoneme models trained following the setting of
  \newcite{vendrov2015order}.}


\subsection{Word boundary prediction}
To explore the sensitivity of the {\sc Phon GRU} model to linguistic structure at the sub-word level, we investigated the encoding of information about word-boundaries in the hidden layers. Logistic regression models were trained on activation patterns of the hidden layers at all timesteps, with the objective of identifying phonemes that preceded a word boundary. For comparison, we also trained logistic regression models on \textit{n}-gram data to perform the same tasks, with positional phoneme \textit{n}-grams in the range $1$-\textit{n}. The location of the word boundaries was taken from the eSpeak transcriptions, which mostly matches the location of word boundaries according to conventional English spelling. However, eSpeak models some coarticulation effects which sometimes leads to word boundaries disappearing from the transcription. For example, {\it bank of a river} is transcribed as \textipa{[baNk @v@ \*rIv@]}.

All models were implemented using the {\tt LogisticRegression} implementation from Scikit-learn \cite{scikit-learn} with L2-regularization. The random samples of 5,000 images each that served as validation and test sets in the visual feature prediction task were used as training and test sets. For the models based on the activation patterns of the hidden layers, the z-score transformation was applied to the activation values to ease optimization. The optimal value of regularization parameter \textit{C} was determined using {\tt GridSearchCV} with 5-fold cross validation on the training set, after which the model with the optimal settings was trained on the full training sample.

Table~\ref{tab:boundary} reports the scores on the test set. The proportion of phonemes preceding a word boundary is 0.29, meaning that predicting {\it no word boundary} by default would be correct in 0.71 of cases. At the highest hidden layer, enough information about the word form is available for correct prediction in 0.82 of cases -- substantially above the majority baseline. The lower levels allow for more accurate prediction of word boundaries: 0.86 at the middle hidden layer, and 0.88 at the bottom level.
Prediction scores of the logistic regression model based on the activation patterns of the lowest hidden layer are comparable to those of a bigram logistic regression model.

These results indicate that information on sub-word structure is only partially encoded by {\sc Phon GRU}, and is mostly absent by the time the signal from the input propagates to the top layer. The bottom layer does learn to encode a fair amount of word boundary information, but the prediction score substantially below 100\% indicates that it is rather selective. 

\begin{table}[h]
    \centering

        \begin{minipage}{0.35\textwidth}
          \centering
          \begin{tabular}{l|rr}
                     Model & Acc @ 5 & Acc @ 10 \\\hline
            {\sc Word Sum} & 0.158     & 0.243 \\
            {\sc Word GRU} & 0.205     & 0.306 \\
            {\sc Phon GRU} & 0.180     & 0.276 \\
          \end{tabular}
          \caption{Image retrieval accuracy at 5 and at 10 on test data for the
            versions of {\sc Word Sum}, {\sc Word GRU} and {\sc Phon
              GRU} chosen by validation.}
          \label{tab:accat5test}
        \end{minipage}
        \hspace{0.5cm}
        \begin{minipage}[r]{0.6\textwidth}
          \centering
          \begin{tabular}{lrrrr}
            \textbf{Model} & & \textbf{Acc} & \textbf{Prec} & \textbf{Rec} \\
            \hline
            Majority & & 0.71 & & \\
            \hline
            Phon GRU & Layer 1 & 0.88 & 0.82 & 0.78 \\
                           & Layer 2 & 0.86 & 0.79 & 0.71 \\
                           & Layer 3 &  0.82 & 0.74 & 0.60 \\
            \hline
            \textit{n}-gram & \textit{n} = 1 & 0.80 & 0.79 & 0.41 \\
                           & \textit{n} = 2 & 0.87 & 0.79 & 0.78 \\
                           & \textit{n} = 3 & 0.93 & 0.86 & 0.90 \\
                           & \textit{n} = 4 & 0.95 & 0.90 & 0.93
          \end{tabular}
          \caption{Prediction scores of logistic regression models based
            on activation vectors of {\sc Phon GRU} and on positional
            \textit{n}-grams}
          \label{tab:boundary}
        \end{minipage}

\end{table}

\subsection{Word similarity}
To understand the encoding of semantic information in {\sc Phon GRU}, we analyzed the cosine similarity of activation vectors for word pairs from the MEN Test Collection \cite{bruni2014multimodal}. The MEN dataset contains 3,000 pairs of English words with semantic similarity judgements on a 50-point scale, which were obtained through crowd-sourcing.\footnote{The MEN dataset is available at \url{http://clic.cimec.unitn.it/~elia.bruni/MEN}}
For each word pair in the MEN dataset, the words were transcribed phonetically using eSpeak and then fed to {\sc Phon GRU} individually. For comparison, the words were also fed to {\sc Word GRU} and {\sc Word Sum}. Word pair similarity was quantified as the cosine similarity between the activation patterns of the hidden layers at the end-of-sentence symbol.
In contrast to {\sc Word GRU} and {\sc Word Sum}, {\sc Phon GRU} has access to the sub-word structure. To explore the role of phonemic form in word similarity, a measure of phonemic difference was included: the Levenshtein distance between the phonetic transcriptions of the two words, normalized by the length of the longer transcription. 

Table~\ref{tab:human} shows Spearman's rank correlation coefficient between human similarity ratings from the MEN dataset and cosine similarity at the last timestep for all hidden layers. In all layers, the cosine similarities between the activation vectors for two words are significantly correlated with human similarity judgements. The strength of the correlation differs considerably between the layers, ranging from 0.09 in the first layer to 0.28 in the highest hidden layer. The second column in Table~\ref{tab:human} shows the correlations when only taking into account the 1283 word pairs of which both words appear at least 100 times in the training set of MS-COCO. 
Correlations for both {\sc Word GRU} and {\sc Word SUM} are considerably higher than for {\sc Phon GRU}. This is expected given that these are word level models with explicit word-embeddings, while {\sc Phon GRU} builds word representations by forwarding phoneme-level input through several layers of processing.

\begin{table}[h]
    \begin{minipage}[l]{0.55\textwidth}
      \centering
      \begin{tabular}{rrr}
        & All words & Frequent words \\\hline
        {\sc Phon GRU} Layer 1 & 0.09 & 0.12\\
        Layer 2 & 0.21 & 0.33 \\
        Layer 3 & 0.28 & 0.45 \\
        \hline
        {\sc Word GRU} & 0.48 & 0.60\\	\hline
        {\sc Word Sum} & 0.42 & 0.56
      \end{tabular}
      \caption{Spearman's correlation coefficient between word-word
        cosine similarity and human similarity judgements. All
        correlations significant at \textit{p} $< 1\mathrm{e}{-4}$.
        Frequent words appear at least 100 times in the training
        data.}
      \label{tab:human}
    \end{minipage}
    \hspace{0.5cm}
    \begin{minipage}[r]{0.4\textwidth}
      \centering
      \begin{tabular}{rr}
        Layer   & $\rho$ \\\hline
        1 & $-0.30$ \\
        2 & $-0.24$ \\
        3 & $-0.15$
      \end{tabular}
      \caption{Spearman's rank correlation coefficient between {\sc
          Phon GRU} cosine similarity and phoneme-level edit
        distance. All correlations significant at \textit{p}
        $< 1\mathrm{e}{-15}$.}
      \label{tab:edit}
    \end{minipage}

\end{table}

Table~\ref{tab:edit} shows Spearman's rank correlation coefficient between the edit distance and the cosine similarity of activation vectors at the hidden layers of {\sc Phon GRU}.
As expected, edit distance and cosine similarity of the activation vectors are negatively correlated: words which are more similar in form are also more similar according to the model.\footnote{Note that in the MEN dataset, meaning and word form are also (weakly) correlated: human similarity judgements and edit distance are correlated at $-0.08$ ($p< 1\mathrm{e}{-5}$).}

The negative correlation between edit distances and cosine similarities is strongest at the lowest hidden layer and weakest, though still present and stronger than for human judgements, at the third hidden layer. 

The correlations of cosine similarities with edit distance on the one hand, and human similarity rating on the other hand, indicate that the different hidden layers reflect increasing levels of representation: whereas at the lowest level mostly encodes information about form, the highest layer mostly encodes semantic information.

\subsection{Position of shared substrings}
Here we quantify the time-scale at which information is retained in
the different layers of {\sc Phon GRU}. We looked at the location of
phoneme strings shared by sentences and their nearest neighbors in the 5,000-image validation sample.
We determined each sentence's nearest neighbor for each hidden layer in {\sc Phon GRU}. The nearest neighbour is the sentence for which the activation vector at the end of sentence symbol has the smallest cosine distance to the activation vector of the original sentence. The position of matching substrings is the average position in the original sentence of symbols in substrings that are shared by the neighbor sentences, counted from the end of the sentence. A high mean average substring position thus means that the shared substring(s) appear early in the sentence. This gives an indirect measure of the timescale at which the different layers operate. Table~\ref{tab:example-shared} shows an example.

As can be seen in Table~\ref{tab:substrings}, the average position of shared substrings in neighbor sentences is closest to the end for the first hidden layer and moves towards the beginning of the sentence for the second and third hidden layer. This indicates a difference between the layers with regards to the timescale they represent. Whereas in the lowest layer only information from the latest timesteps is present, the higher layers retain the input signal over longer timescales.

\begin{table}[h]
    \begin{minipage}{0.6\textwidth}
      \begin{tabular}{c}
        Layer 1 \\\hline
        A metallic bench {\bf on a path in} the {\bf park} \\
        A man riding a bicycle {\bf on a path} in a {\bf park} \\\hline
        Layer 3 \\\hline
        A metallic {\bf bench} on a path in the {\bf park} \\
        A stone park {\bf bench} sitting in an empty green {\bf park}\\ \hline
      \end{tabular}
      \caption{An illustrative sentence with its nearest neighbour at
        layer 1 and layer 3. For readability, sentences are displayed
        in conventional spelling, and only highlight matching
        substrings of length $\geq3$. In reality we used phonetic
        transcriptions to compute shared substring positions, and 
        substrings of all lengths. }
      \label{tab:example-shared}
    \end{minipage}
    \hspace{0.5cm}
    \begin{minipage}{0.35\textwidth}
      \centering
      \begin{tabular}{rr}
        Layer   & Mean position \\\hline
        1 & 12.1 \\
        2 & 14.9 \\
        3 & 16.8 \\
      \end{tabular}
      \caption{Average position of phonemes in shared substrings
        between nearest neighbour sentences according to {\sc Phon
          GRU} representations at the different layers. Positions are
        indexed from end of string.}
      \label{tab:substrings}
    \end{minipage}

\end{table}

\section{Future work}
Although our analyses show a clear pattern of short-timescale
information in the lower layers and larger dependencies in the higher
layers, the third layer still encodes information about the phonetic
form: its activation patterns were predictive of word boundaries, and
similarities between word pairs at this level were more strongly
correlated with edit distance than human similarity judgements are. It
would be interesting to investigate exactly what information that is,
and to what extent it is analogous to language representation in the
mind of human speakers. In humans both word phonological form and word
meaning can act as primes, which is somewhat reminiscent of the
behavior of  our model. 

Finally, we would like to take the next step towards grounded learning of language from raw perceptual input, and apply models similar to the one described here to acoustic speech signal coupled with visual input. We expect this to be a challenging but essential endeavor.


\bibliographystyle{emnlp2016}
\bibliography{biblio}

\end{document}